\documentclass[letterpaper]{article}
\usepackage{aaai2027}
\usepackage[hyphens]{url}
\usepackage{graphicx}
\usepackage{natbib}
\usepackage{caption}
\usepackage{booktabs}
\title{How Much Should an Agent Remember?\\Deployment Lessons from Self-Distilled Memory for LLM Agents}
\author{K.~R. Jayaram, Vatche Isahagian, Aswath Vaithinathan Aravindan, Vinod Muthusamy,\\
Punleuk Oum, Gaodan Fang, Michal Ulewicz, Gegi Thomas, and Evelyn Duesterwald}
\affiliations{IBM}

\begin{document}
\maketitle

\begin{abstract}
LLM agents on long-horizon tasks fail in recurring, procedural ways. They mis-paginate an API, resolve the wrong entity, or answer when no answer was asked. We built a pipeline that lets an agent learn from its own history. It first mines short \emph{lessons} from the agent's own trajectories, one to three per episode. It then consolidates those lessons into a smaller set of \emph{guidelines}, where each guideline records how many lessons it merged. At inference the agent receives either all guidelines or a selected subset. The pipeline ships in the open-source altk-evolve library and is deployed in IBM's CUGA agent and watsonx Orchestrate platform. We report what we learned building, measuring, and productizing it. Our primary instrument is the AppWorld benchmark, across models spanning 40--92\% baseline task completion. The central finding is that the right amount of memory is a property of the model, not the memory. A strong model keeps improving as guidelines are added. DeepSeek-V3.2 gains +9.5 points of task completion and +16.1 points of scenario completion with its full 191-guideline set. A weak model gains less from the same full set than from a small selected subset. gpt-oss-120b starts at 39.9\% task completion. It reaches 51.2\% with all 199 of its guidelines but 55.4\% (mean of two runs) with a selected subset of about 29. Around that result we report what went wrong: a high-support guideline that backfired, provenance silently dropped during consolidation, a swallowed API error that failed every task while producing a plausible log, and a baseline confound that nearly produced a misleading comparison with a prior method. We distill these into recommendations for teams deploying agent memory.
\end{abstract}

\section{Introduction}

Agents built on large language models fail on realistic multi-step tasks in a distinctive way. We read hundreds of trajectories of IBM's CUGA agent and of the AppWorld reference agent. The failures are rarely knowledge gaps. The model knows the API, yet it mis-paginates it. It knows the user has contacts, yet it resolves the wrong one. It knows how to return an answer, yet it returns one when nothing was asked. These are procedural lapses, and they recur. That makes them look learnable from the agent's own history.

That observation is now widely shared. Recent systems let an agent write notes about its experience and read them back later. Examples include ExpeL \citep{zhao2024expel}, Agentic Context Engineering (ACE) \citep{zhang2025ace}, and reflective memory in generative agents \citep{park2023generative}. In products, Anthropic's ``dreaming'' consolidates memory for managed agents \citep{anthropic2026dreams}. What remains scarce is an account of what happens when such a system is built, measured carefully, and shipped.

This paper is that account. Our pipeline has three stages. It first extracts \emph{lessons} from the agent's own trajectories: short, imperative notes about what to do differently, one to three per episode. It then consolidates those lessons into \emph{guidelines}, a smaller set in which each guideline records how many lessons support it. Finally, it delivers guidelines to the agent at inference, either all of them or a selected subset. The pipeline is part of the open-source altk-evolve library \citep{altkevolve2026}. It is deployed in IBM products, including the CUGA agent and the watsonx Orchestrate platform. To measure it we used AppWorld \citep{trivedi2024appworld}. AppWorld offers three things a deployment needs and a production log rarely provides: an authoritative pass/fail verdict per task, graded difficulty, and a clean train/dev/test separation.

The method is not the contribution of this paper. Several of its pieces have precedents, which we credit below. The contribution is what we learned deploying it across models from 40\% to 92\% baseline task completion. In the spirit of this track, we focus on what went wrong:
\begin{itemize}\setlength{\itemsep}{1pt}
\item The right amount of memory is a property of the model. A strong model keeps improving as guidelines are added. A weak model gains more from a small, selected subset than from the same full set (Insight~1).
\item Compressing memory to a tidy checklist discards most of its value. The value lives in one-off, situational lessons (Insight~2).
\item Selection is only possible if provenance survives consolidation. During productization it silently did not, twice (Insight~3).
\item A frequently derived guideline can still be wrong. Support measures recurrence, not correctness (Insight~4).
\item Run-to-run noise on a fixed configuration is large enough to fabricate a ``win'' (Insight~5).
\item Comparing memory systems across different agent scaffolds nearly produced a misleading headline (Insight~6).
\item Silent failures cost more calendar time than any modeling problem: a swallowed HTTP error, an idle-dropped connection, and a write hook that never fired (Insight~7).
\end{itemize}
We close with the practices we now apply and with a section on deployment in IBM's products. That section reports CUGA benchmark results re-collected after the write-path bug described in Insight~7, with GPT-4.1 and GLM-5.3-Flash, and a qualitative example from watsonx Orchestrate.

\section{The System: altk-evolve}

\paragraph{The library.}
altk-evolve is an open-source library for agents that learn from their own runs \citep{altkevolve2026}. Its guideline generation builds on our group's earlier work on trajectory-informed memory \citep{fang2026trajectory}. It captures trajectories in three ways. It can instrument common client libraries directly (OpenAI, LiteLLM, smolagents, and the OpenAI Agents SDK). It can sync traces from an observability store such as Arize Phoenix. Or an agent can hand over a trajectory through an MCP tool at the end of a task. From each trajectory an LLM extracts short lessons, optionally after segmenting the trajectory into subtasks. Lessons, and the guidelines consolidated from them, are stored as entities with metadata (category, rationale, trigger, source task, and support) in a pluggable backend: a filesystem store for development, or Milvus or Postgres with pgvector for semantic search. Consolidation clusters related lessons on their source-task embeddings and merges within each cluster with an LLM. It is support-conserving: every guideline records how many lessons it merged, and delivery can select by that count. Both are described below. Agents consume memory through an MCP server that exposes tools such as \texttt{get\_guidelines}, \texttt{get\_relevant\_guidelines}, and \texttt{save\_trajectory}. A Python client, a CLI, and a web UI serve developers and reviewers. CUGA connects to the library over MCP. The rest of this section describes each stage as we ran it for this study.

\paragraph{Extraction.}
We run the agent on the benchmark's train and dev splits and keep every trajectory, successes and failures alike. For each trajectory, the extraction stage prompts an LLM to write one to three short imperative lessons about what to do differently next time. Whether a lesson counts as success- or failure-evidence comes from the task's actual outcome, as judged by the evaluator. It never comes from the LLM's self-report. For DeepSeek-V3.2, 147 train/dev trajectories yield 290 raw lessons. Of these, 62 come from failures and 228 from successes.

\paragraph{Why consolidation is forced.}
Raw episodic memory does not stay small. Most of what accumulates is near-duplicate. Five different tasks discover pagination the hard way, and all five write it down. Left alone, this breaks the natural delivery mechanism, retrieval. The top-$k$ slots fill with copies of the same lesson. Genuinely different guidance is crowded out. Irrelevant one-offs arrive as dead context. Consolidation is therefore not a cost optimization. It is what makes memory retrievable at all.

\paragraph{Support-conserving consolidation.}
The consolidation stage embeds the lessons and clusters near-duplicates. It then prompts an LLM to merge only within a cluster, and only when following one lesson is equivalent to following the other. Each surviving guideline carries a \emph{support} count: how many independent lessons collapsed into it. It also carries a merged evidence polarity (success, failure, or both). One invariant is enforced: the sum of support over the output equals the number of raw lessons. Anything the LLM fails to cover is carried through unchanged. For DeepSeek this turns 290 lessons into 191 guidelines, of which 144 have support one.

\paragraph{Delivery.}
A guideline set can be injected statically, or delivered by retrieval. In retrieval, a \emph{core} of high-support guidelines (support $\geq 3$) is always included. The top-$k$ further guidelines are then added, chosen by how similar their \emph{source task} is to the current task. Candidates already covered by the core, or duplicating one another, are dropped. Thresholding by support (``sup2'', ``sup3'') happens at selection time and never deletes anything from the store. In altk-evolve these are configuration: \texttt{consolidation\_mode}, \texttt{min\_support}, \texttt{core\_support}, \texttt{retrieval\_top\_k}, and \texttt{injection\_mode}.

\paragraph{Deployment constraint.}
altk-evolve generally uses the same model that the agent uses. In this study, the model that ran the agent is also the model that extracted its lessons and consolidated them. There is no separate teacher model and no human labeling. This mirrors the deployment we care about, an agent that improves from its own logs. It also makes the capability-dependence in Insight~1 an unavoidable design question rather than a curiosity.

\section{Measurement Setting}

AppWorld simulates nine applications (email, payments, music, shopping, and others) behind 457 APIs and about a hundred simulated people \citep{trivedi2024appworld}. Its 750 tasks require writing and executing Python against those APIs. We use \texttt{test\_normal} (168 tasks) throughout and \texttt{test\_challenge} (417 tasks) where noted. The agent is the reference ReAct code agent. Each step is one model call that writes Python, and the environment returns the output. We report two metrics. TGC is the fraction of tasks that pass all of their evaluation checks. SGC is the stricter fraction of scenarios in which every task variant passes. Eight models were baselined and six were run with memory. All memory is mined from train/dev only. Because dev is consumed by mining, no held-out split was left for choosing a configuration. We compared configurations directly on \texttt{test\_normal}: five for DeepSeek-V3.2, four for gpt-oss-120b, two each for GPT-5.5, GLM-5, and Gemma-4-31B, and one for Claude Opus 4.6. Table~\ref{tab:models} therefore reports, next to each model's best configuration, a fixed rule decided before any test run: inject all guidelines. Results are single runs, the benchmark's convention, except where noted. Insight~5 quantifies what that hides.

\begin{figure*}[t]
\centering
\includegraphics[width=\textwidth]{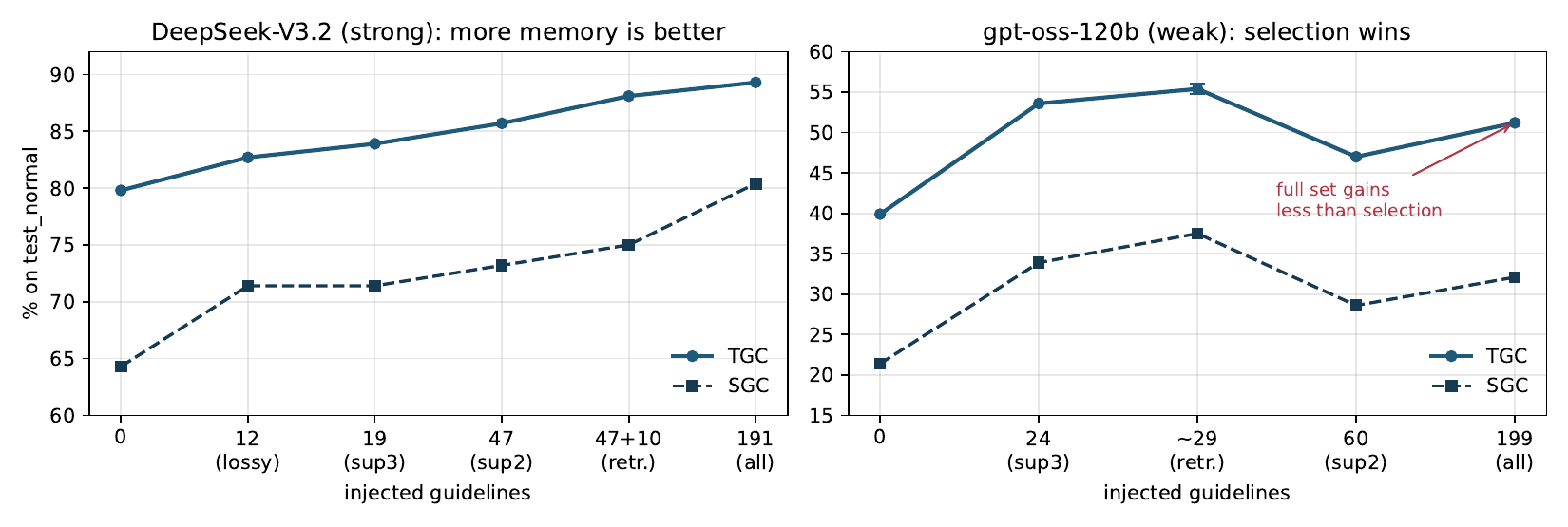}
\caption{Dose--response on \texttt{test\_normal}. Left: a strong model keeps improving as the injected guideline set grows from 0 to all 191 (``retr.'' = 47 high-support guidelines plus 10 retrieved per task). Adjacent points differ by two to four tasks, within noise. The endpoints differ by 16. Right: a weak model gains most from about 29 selected guidelines (mean of two runs, bar shows both). All 199 are no better than a 24-guideline core.}
\label{fig:dosage}
\end{figure*}

\section{Insights}

\subsection{Insight 1: The right dose is a property of the model}

\emph{What we expected.} That memory would help every model roughly in proportion to its headroom.

\emph{What happened.} Table~\ref{tab:models} and Figure~\ref{fig:dosage}. Three regimes appeared. A strong model with headroom keeps improving as memory grows. DeepSeek-V3.2 climbs 79.8 $\rightarrow$ 82.7 $\rightarrow$ 83.9 $\rightarrow$ 85.7 $\rightarrow$ 89.3 TGC as the injected set grows from 0 to 12 to 19 to 47 to all 191 guidelines. Each step is two to four tasks and within noise on its own. The trend is not: the endpoints are 16 tasks apart. SGC rises from 64.3 to 80.4. The gain holds on the hard split (60.4 $\rightarrow$ 72.9 TGC, 41.7 $\rightarrow$ 58.3 SGC). GPT-5.5, already at 92.3, still gains with all 207 of its own guidelines (95.2/89.3). The TGC gain is about five tasks, at the noise floor. The SGC gain is larger. A weak model shows a different curve. gpt-oss-120b starts at 39.9. With all 199 guidelines it scores 51.2. That is no better than the 53.6 it scores with a 24-guideline core, a gap of four tasks. Its best result comes from about 29 selected guidelines: 55.4 TGC, the mean of two runs (56.0 and 54.8). That is seven tasks above the full set, at essentially baseline token cost (116K vs.\ 110K per task). The fixed all-guidelines rule in Table~\ref{tab:models} leaves every model at or above its baseline and keeps both large gains: +9.5 for DeepSeek and +11.3 for gpt-oss. Two further regimes bracket the range. GLM-5 is saturated: it scores 87.5 with or without its 72 guidelines. Gemma-4-31B is too weak to apply memory: a wash overall, and hurt by an over-tight core.

\emph{What we changed.} The first implementation had one delivery mode. The shipped one exposes dose as configuration. It defaults conservatively to retrieval with a high-support core, because the failure mode of too much memory on a weak model is silent. The agent does not error. It just does worse. We also stopped saying ``memory helps.'' The honest statement is that the optimal amount is a function of capability. A memory system that cannot vary it will be wrong for some of its users.

\paragraph{Where the gains land.}
Table~\ref{tab:diff} splits the same runs by task difficulty. Two patterns held across models. For capable models the \emph{middle} tier benefits most. DeepSeek's medium-difficulty SGC jumps from 62.5 to 93.8, a 31-point swing, because guidelines flip borderline cases that were almost right. For a weak model the hard tier is a ceiling that no guideline moves: gpt-oss hard-tier SGC stays near the floor at 9.5. A frontier model does move the hard tier on the hard split (Opus 4.6, 73.3 $\rightarrow$ 83.1 TGC on \texttt{test\_challenge}). For a deploying team this is the ROI map. Expect memory to convert near-misses, not to lift a model over tasks it cannot do at all.

\begin{table}[t]
\centering\scriptsize
\setlength{\tabcolsep}{3pt}
\begin{tabular}{@{}llccc@{}}
\toprule
Model (split) & Metric & Easy & Medium & Hard \\
\midrule
DeepSeek-V3.2 (normal) & TGC & 93.0$\rightarrow$94.7 & 81.2$\rightarrow$97.9 & 66.7$\rightarrow$77.8 \\
 & SGC & 84.2$\rightarrow$84.2 & 62.5$\rightarrow$93.8 & 47.6$\rightarrow$66.7 \\
gpt-oss-120b (normal) & TGC & 66.7$\rightarrow$82.5 & 35.4$\rightarrow$56.2 & 19.1$\rightarrow$31.8 \\
 & SGC & 42.1$\rightarrow$57.9 & 18.8$\rightarrow$50.0 & 4.8$\rightarrow$9.5 \\
Claude Opus 4.6 (challenge) & TGC & 73.6$\rightarrow$95.8 & 72.0$\rightarrow$78.0 & 73.3$\rightarrow$83.1 \\
 & SGC & 66.7$\rightarrow$91.7 & 58.0$\rightarrow$70.0 & 61.5$\rightarrow$72.3 \\
\bottomrule
\end{tabular}
\caption{Baseline $\rightarrow$ best memory configuration, by difficulty tier. For gpt-oss this is the first of its two retrieval runs (56.0 TGC). Capable models gain most on the middle tier. The hard tier is a ceiling for the weak model.}
\label{tab:diff}
\end{table}

\begin{table}[t]
\centering\scriptsize
\setlength{\tabcolsep}{2pt}
\begin{tabular}{@{}lrrrlrr@{}}
\toprule
 & & \multicolumn{2}{c}{Fixed: all} & \multicolumn{3}{c}{Best on test} \\
\cmidrule(lr){3-4}\cmidrule(l){5-7}
Model & Base & TGC & $\Delta$ & Config (\#) & TGC/SGC & $\Delta$ \\
\midrule
GPT-5.5 & 92.3 & 95.2 & +2.9 & all (207) & 95.2/89.3 & +2.9 \\
Opus 4.6 & 90.5 & -- & -- & lossy (12)$^\dagger$ & 94.6/94.6 & +4.1 \\
GLM-5 & 87.5 & 87.5 & 0.0 & all (72) & 87.5/80.4 & 0.0 \\
DeepSeek-V3.2 & 79.8 & 89.3 & +9.5 & all (191) & 89.3/80.4 & +9.5 \\
Gemma-4-31B$^\ast$ & 63.4 & 64.6 & +1.2 & all (87) & 64.6/45.5 & +1.2 \\
gpt-oss-120b & 39.9 & 51.2 & +11.3 & retr.\ ($\approx$29)$^\ddagger$ & 55.4/37.5 & +15.5 \\
\bottomrule
\end{tabular}
\caption{Memory per model on \texttt{test\_normal}. Each model mined its own memory. ``Fixed: all'' injects every guideline, a rule set before any test run. ``Best on test'' was chosen among the configurations run on \texttt{test\_normal}. $^\dagger$Only the 12-guideline compressed set was run for Opus. $^\ast$On the 164-task subset its memory runs used (62.5 baseline on all 168). $^\ddagger$TGC is the mean of two runs (56.0, 54.8). SGC is from the first.}
\label{tab:models}
\end{table}

\subsection{Insight 2: Do not compress. The value is in the long tail}

\emph{What we expected.} Our first consolidation prompt asked the LLM for a small set of highly relevant rules, which is what a person would ask for. It produced twelve, and they read well.

\emph{What happened.} Those twelve gave DeepSeek +2.9 TGC. The full set of 191 gave +9.5 and more than doubled the SGC gain. The two differ by eleven tasks, well above noise. Every intermediate dose fell between them. The difference is the 144 guidelines with support one. These are situational lessons that a summarizer discards as unimportant, precisely because each occurred once. This corroborates, from outside, the ``brevity bias'' and ``context collapse'' failure modes that the ACE authors describe \citep{zhang2025ace}.

\emph{What we changed.} Lossless became the default consolidation mode, and compression became an explicit opt-in. Insight~1 is the necessary qualifier. On a weak model the full long tail is exactly what drowns it. ``Keep everything'' is only right when delivery can select.

\subsection{Insight 3: Selection needs provenance, and provenance is easy to lose}

\emph{What we expected.} That consolidation and selection were independent stages.

\emph{What happened.} They are not. Selection by support requires that support survive consolidation. Retrieval by source task requires that the source task survive it. When we productized into altk-evolve we found two independent ways this had already broken. An earlier version of the consolidation stage rebuilt each merged guideline's metadata from a fixed field list. It silently dropped any provenance not on the list, including the support count and a generation-method tag another team was relying on. Separately, the first version of support attribution double-counted a member whenever the LLM repeated an index inside one merge. An automated code reviewer caught it before a person did.

\emph{What we changed.} Conservation is now an enforced invariant with a regression test, not a convention. Consolidation reports total support before and after. Any lesson the LLM fails to cover is carried through verbatim. The lesson generalizes to any memory pipeline. If a later stage, such as selection, filters or ranks on a field, then every earlier stage, such as extraction and consolidation, must preserve that field as an invariant.

\subsection{Insight 4: Support is recurrence, not correctness}

\emph{What we expected.} That a lesson derived independently by many episodes was safe to trust.

\emph{What happened.} A high-support pagination guideline read ``loop until fewer than the page limit are returned.'' The rule is correct only if the agent knows the page limit, and it did not. It assumed a limit of 20, while the API's default page size was 5. The first page therefore held 5 items. Since 5 is fewer than 20, the agent concluded that it had seen the last page and stopped. It missed every later page. That was a regression on a task the baseline had passed. We fixed it by hand. The reworded guideline says ``pass an explicit page limit and loop until a page comes back empty,'' which no longer depends on guessing the default. The task recovered, and hard-tier TGC rose from 66.7 to 73.0.

\emph{What we changed.} Nothing automatic, and that is the insight. Support counts how often a lesson was derived. That is evidence of generality, not of correctness. A popular wrong lesson scores high. Our pipeline has no notion of a guideline being contradicted, stale, or harmful. Systems designed as ongoing maintenance loops do. Anthropic's dreaming resolves contradictions in favor of the most recent value and prunes stale entries \citep{anthropic2026dreams}. ExpeL lets the model down-vote an insight to deletion \citep{zhao2024expel}. Generative agents weight retrieval by recency \citep{park2023generative}. We now regard a correctness or recency signal as the most important missing component. We would not run an unattended memory loop without one.

\subsection{Insight 5: Measure the noise floor before believing a delta}

\emph{What we expected.} That temperature zero and a fixed seed made runs comparable.

\emph{What happened.} Two runs of an identical gpt-oss-120b configuration scored 56.0 and 54.8 TGC. Across identical DeepSeek runs, roughly 32 of 168 tasks flipped outcome in one direction or the other while the aggregate barely moved. Sibling tasks in the same scenario flipped in opposite directions on unrelated prompt edits. Embedding similarity between a task and its retrieved guidelines did not predict which tasks improved. Improved and regressed tasks had mean retrieval cosine of 0.463 and 0.468.

\emph{What we changed.} We repeat runs wherever a claim rests on a small gap. We treat differences under about five tasks as noise. Concretely, this turned a ``+1.2 TGC win over ACE'' on gpt-oss into what it is: a tie on accuracy at one-seventh the token cost. That is both more defensible and, for a deploying team, more useful.

\subsection{Insight 6: Compare on a fixed scaffold, and report tokens}

\emph{What we expected.} That two memory methods evaluated on the same benchmark and model were comparable on their final scores.

\emph{What happened.} Table~\ref{tab:ace}. On DeepSeek our system reached 89.3/80.4 against ACE's 80.4/73.3. But the two agents' no-memory baselines were 79.8 and 72.0. ACE's ReAct variant carries a different prompt template, with a playbook slot and a longer set of worked examples. Measured as lift over its own baseline, ACE gains +8.4/+16.2 and ours +9.5/+16.1. The memory contributions are comparable, and the headline gap is mostly scaffold. What survives scrutiny is different from what we first wrote. On DeepSeek we get same-or-better accuracy at roughly 40\% of ACE's tokens (263K vs.\ 634K per task). On the weak model we get a tie at one-seventh the tokens (116K vs.\ 777K). There, ACE's whole-playbook injection scores 54.8 and our own full set scores 51.2. The win is selection, not the memory itself.

\emph{What we changed.} We hold the scaffold fixed and vary only the memory when comparing methods. We report per-task tokens alongside accuracy in every table, and we lead with deltas. The token column turned out to be the most decision-relevant number for the deploying team.

\paragraph{Cost.}
The overhead is almost entirely input-side, because the injected set is re-sent on every reasoning step. Output tokens are flat, and steps barely move (DeepSeek, 18.4 $\rightarrow$ 19.3 per task). The full 191-guideline set costs 78\% more tokens per task than baseline. The 47-guideline support-$\geq$2 set is the efficiency sweet spot, at +27\% tokens for +5.9 TGC. For the weak model, retrieval keeps cost within about 5\% of baseline, while the full set inflates it by half. Caching the static prefix, or retrieving, is the lever.

\begin{table}[t]
\centering\footnotesize
\setlength{\tabcolsep}{4pt}
\begin{tabular}{@{}llrrr@{}}
\toprule
System & Guidelines & TGC & SGC & Tok/task \\
\midrule
\multicolumn{5}{@{}l}{\emph{DeepSeek-V3.2}}\\
ACE-ReAct, no memory & 0 & 72.0 & 57.1 & 221K \\
ACE-ReAct + ACE & 106 & 80.4 & 73.3 & 634K \\
ReAct, no memory & 0 & 79.8 & 64.3 & 148K \\
ReAct + ours (all) & 191 & \textbf{89.3} & \textbf{80.4} & 263K \\
\midrule
\multicolumn{5}{@{}l}{\emph{gpt-oss-120b}}\\
ReAct, no memory & 0 & 39.9 & 21.4 & 110K \\
ACE-ReAct + ACE & full & 54.8 & 35.7 & 777K \\
ReAct + ours (all) & 199 & 51.2 & 32.1 & 166K \\
ReAct + ours (retrieval) & $\approx$29 & \textbf{55.4}$^\ddagger$ & \textbf{37.5} & 116K \\
\bottomrule
\end{tabular}
\caption{Comparison with ACE on \texttt{test\_normal}. ACE was run in-house on the same models and splits. Note the differing no-memory baselines on DeepSeek (Insight~6). $^\ddagger$Mean of two runs (56.0, 54.8). SGC is from the first.}
\label{tab:ace}
\end{table}

\subsection{Insight 7: Silent failures cost more than modeling problems}

Three operational failures consumed more calendar time than any research question. First, a GPT-5-class model rejects a temperature of zero with an HTTP 400. The harness caught and swallowed the error. Every task ``failed'' after exactly one model call, and the run completed with a plausible-looking log. Second, consolidating hundreds of lessons in a single call exhausted the output budget on reasoning models, which returned truncated JSON. Chunked map-reduce to a fixed point fixed it. Third, multi-hour evaluations through a corporate proxy died whenever the network went idle, stranding jobs. In the CUGA integration, memory \emph{writes} were never invoked on one code path. The store stayed empty, and memory was silently read-only from cold start. None of these produced an error a person saw. Our recommendations are unglamorous. Fail loudly on any per-task exception. Assert a minimum number of model calls per task. Watchdog progress on long runs. Check that the memory store's entity count actually grows.

\section{Related Work}

Learning from one's own trajectories has a substantial lineage. Reflexion stores verbal self-reflections per episode \citep{shinn2023reflexion}. ExpeL extracts cross-task natural-language insights from training tasks \citep{zhao2024expel}. It maintains them with ADD, EDIT, UPVOTE, and DOWNVOTE operations and an importance count that removes an insight at zero. That count is a model's vote on an insight's quality. Our support is instead a count of independent derivations, conserved through merging. ExpeL also injects all insights at inference and retrieves only past trajectories by similarity. Agent Workflow Memory induces reusable workflows from trajectories \citep{wang2024awm}. ACE grows a playbook through generation, reflection, and curation with incremental updates, explicitly to avoid brevity bias and context collapse \citep{zhang2025ace}. Insight~2 corroborates that concern. Insight~1 adds the case ACE does not study: the weak model that must be given less. Generative agents periodically reflect over a memory stream and retrieve by recency, importance, and relevance \citep{park2023generative}. Our pipeline has the latter two and, as Insight~4 notes, lacks recency. Offline consolidation itself is older. It appears in the wake-sleep algorithm \citep{hinton1995wakesleep}, in DreamCoder's sleep phase that refactors solved programs into a reusable library \citep{ellis2021dreamcoder}, and in sleep-time compute for agents \citep{lin2025sleeptime,packer2023memgpt}. Anthropic's dreaming is the same idea productized. It resolves contradictions by recency but, as far as its documentation shows, preserves no multiplicity \citep{anthropic2026dreams}. All of these, and our episodic-to-semantic framing, descend from complementary learning systems theory \citep{mcclelland1995cls}.

\section{Deployment in IBM's Products}

\paragraph{CUGA.}
CUGA is IBM's multi-agent assistant. Memory was integrated into three of its sub-agents: task analysis, API shortlisting, and task decomposition. Each receives agent-specific tips retrieved by task intent. An initial 50-task evaluation found no statistically significant effect on pass rate or step count. We later found that it had run with the write-path bug of Insight~7, so the memory store never grew past its seed.

With the write path repaired, we re-ran the CUGA integration with two models. With GPT-4.1 we used two benchmarks and averaged over three trials. The first is AppWorld (Table~\ref{tab:cuga-appworld}). The second is the retail split of $\tau$-bench \citep{yao2024tau, barres2025tau2}, a customer-service suite of 74 training and 40 test tasks (Table~\ref{tab:cuga-taubench}). On AppWorld, lossless consolidation raises task completion from 71.03\% to 72.42\% and scenario completion from 54.76\% to 58.93\%. Steps per task fall slightly, from 9.63 to 9.31. On $\tau$-bench, our sup3 configuration (12 guidelines) lifts mean accuracy from 51.7\% to 55.8\%. It also lifts the stricter pass\textasciicircum{}3 measure, which credits a task only when all three of its trials pass, from 27.5\% to 32.5\%, at marginally fewer conversational turns per task. Each of these deltas is about two tasks per trial. That is below the noise floor of Insight~5. They show that the repaired integration works end to end and that its effect points the same way on two benchmarks. They do not establish its size.

To measure size we needed a model with a stronger signal. We repeated the AppWorld experiment with GLM-5.3-Flash, an open-weight model that mined its own memory from CUGA's train and dev runs: 280 lessons, consolidated losslessly into 164 guidelines. Table~\ref{tab:cuga-glm} reports two runs per arm on \texttt{test\_normal}. The full set raises TGC from 87.8 to 93.2, about nine tasks, and SGC from 84.8 to 88.4. Both memory runs score above both no-memory runs. Steps per task do not change. The cost is 36\% more tokens per task. The 29-guideline sup3 subset stays within noise of the baseline. This is the strong-model pattern of Insight~1 in a second scaffold: the full set helps and a compressed one does not.

\begin{table}[t]
\centering\footnotesize
\setlength{\tabcolsep}{4pt}
\begin{tabular}{@{}lccc@{}}
\toprule
Method & TGC & SGC & Steps/Task \\
\midrule
Base CUGA & 71.03 & 54.76 & 9.63 \\
Ours (lossless) & 72.42 & 58.93 & 9.31 \\
$\Delta$ vs.\ base & +1.39 & +4.17 & $-$0.32 \\
\bottomrule
\end{tabular}
\caption{CUGA with the GPT-4.1 model on AppWorld, mean over three trials. Lossless memory improves both completion metrics (TGC over 168 tasks, SGC over scenarios) while slightly reducing steps per task. Base CUGA passes 119.3 tasks and 30.7 scenarios on average. Lossless passes 121.7 and 33.0.}
\label{tab:cuga-appworld}
\end{table}

\begin{table}[t]
\centering\footnotesize
\setlength{\tabcolsep}{4pt}
\begin{tabular}{@{}lrrr@{}}
\toprule
Metric & Baseline & Ours (sup3) & $\Delta$ \\
\midrule
Mean accuracy & 51.7\% & 55.8\% & +4.1 \\
pass\textasciicircum{}3 (all 3 trials pass) & 27.5\% & 32.5\% & +5.0 \\
Avg turns / task & 31.3 & 30.8 & $-$0.5 \\
\bottomrule
\end{tabular}
\caption{CUGA with the GPT-4.1 model on the $\tau$-bench retail split (40 test tasks), mean over three trials. Ours here is the sup3 configuration with 12 guidelines. It improves both mean accuracy and the stricter pass\textasciicircum{}3 measure (which credits a task only when all three of its trials pass) at slightly fewer turns per task.}
\label{tab:cuga-taubench}
\end{table}

\begin{table}[t]
\centering\footnotesize
\setlength{\tabcolsep}{4pt}
\begin{tabular}{@{}lrccrr@{}}
\toprule
Arm & Guid. & TGC & SGC & Steps & Tok/task \\
\midrule
No memory & 0 & 87.8 & 84.8 & 33.8 & 219K \\
sup3 & 29 & 88.4 & 83.9 & 33.8 & 234K \\
Lossless & 164 & \textbf{93.2} & \textbf{88.4} & 33.3 & 297K \\
\bottomrule
\end{tabular}
\caption{CUGA with GLM-5.3-Flash on AppWorld \texttt{test\_normal}, mean of two runs per arm. TGC ranges: no memory 87.5--88.1, sup3 87.5--89.3, lossless 92.9--93.5. The model mined its own guidelines from CUGA's train and dev runs.}
\label{tab:cuga-glm}
\end{table}

\paragraph{watsonx Orchestrate.}
The same memory library is also integrated into IBM's watsonx Orchestrate,\footnote{\url{https://www.ibm.com/products/watsonx-orchestrate}} the company's enterprise platform for building and running assistants that automate business tasks. Any agent built on the platform can learn from its own prior runs. A qualitative production trace illustrates cross-task transfer. On a source expense-claim task for employee E-1042, before the assistant had learned a relevant guideline, it attempted submission before converting the HEIC receipt. It invoked \texttt{submit\_expense\_claim} twice (13 steps, 18.53\,s). Repeated tool calls are not merely wasteful when they have expensive or irreversible side effects. In finance or insurance workflows, for example, they can create duplicate claims or transactions. A procedural guideline distilled from that trajectory was later retrieved for a different task: employee E-2048 with a TIFF receipt. The assistant transferred the learned procedure across task instances. It converted the receipt to PDF via \texttt{convert\_expense\_attachment\_to\_pdf} before invoking the submission tool once (10 steps, 11.35\,s), avoiding the repeated submission call and reducing observed latency by approximately 39\%. Because the inputs differ, the step and latency values are descriptive rather than a controlled estimate of improvement. The evidence here is the transfer of the learned procedure to a new task. Figures~\ref{fig:wxo-chat} and \ref{fig:wxo-timeline} document the source and transfer traces. This is a single qualitative example rather than a benchmark. But it is exactly the kind of recurring, learnable procedural lapse that motivated the system (Introduction), learned from one task and corrected in another in a shipped product.

\section{Conclusion and Recommendations}

Memory distilled from an agent's own history is worth deploying. With nothing but its own train-time trajectories, a mid-tier model gained +9.5 TGC and +16.1 SGC. A weak one gained +11.3 TGC with all of its guidelines and +15.5 with a selected subset. But the value did not come from the idea. It came from what we learned deploying it. For teams about to do the same:
\begin{enumerate}\setlength{\itemsep}{1pt}
\item Treat dose as a first-class parameter and measure it per model. Assume nothing about monotonicity.
\item Consolidate losslessly and select at delivery. Do not compress by default.
\item Make every field a downstream stage ranks on an enforced, tested invariant of every upstream stage.
\item Add a correctness or recency signal before running memory unattended. Recurrence is not truth.
\item Establish the noise floor with repeated runs before reading any small delta.
\item Compare memory systems on a fixed scaffold, lead with deltas, and always report tokens.
\item Assume failures will be silent. Assert on call counts and store growth, and fail loudly.
\end{enumerate}

\bibliography{refs}

\appendix
\section{Appendix: watsonx Orchestrate Deployment}

\begin{figure*}[htb]
\centering
\includegraphics[width=\textwidth]{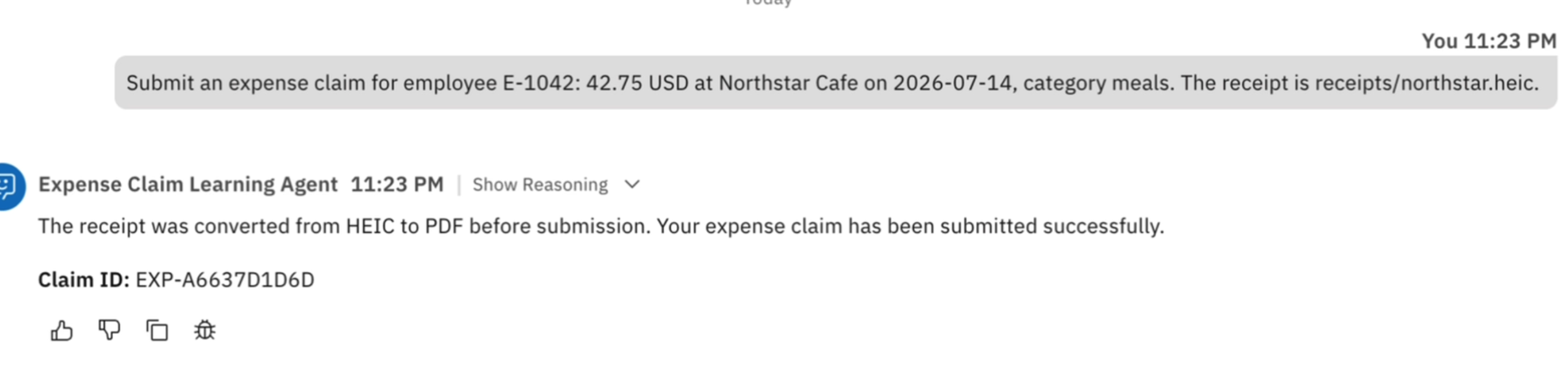}
\caption{Source task in watsonx Orchestrate (E-1042). Its trajectory yielded a guideline to convert unsupported receipts before submission.}
\label{fig:wxo-chat}
\end{figure*}

\begin{figure*}[htb]
\centering
\textbf{Source task (E-1042, HEIC): source trajectory}\\[3pt]
\includegraphics[width=0.58\textwidth]{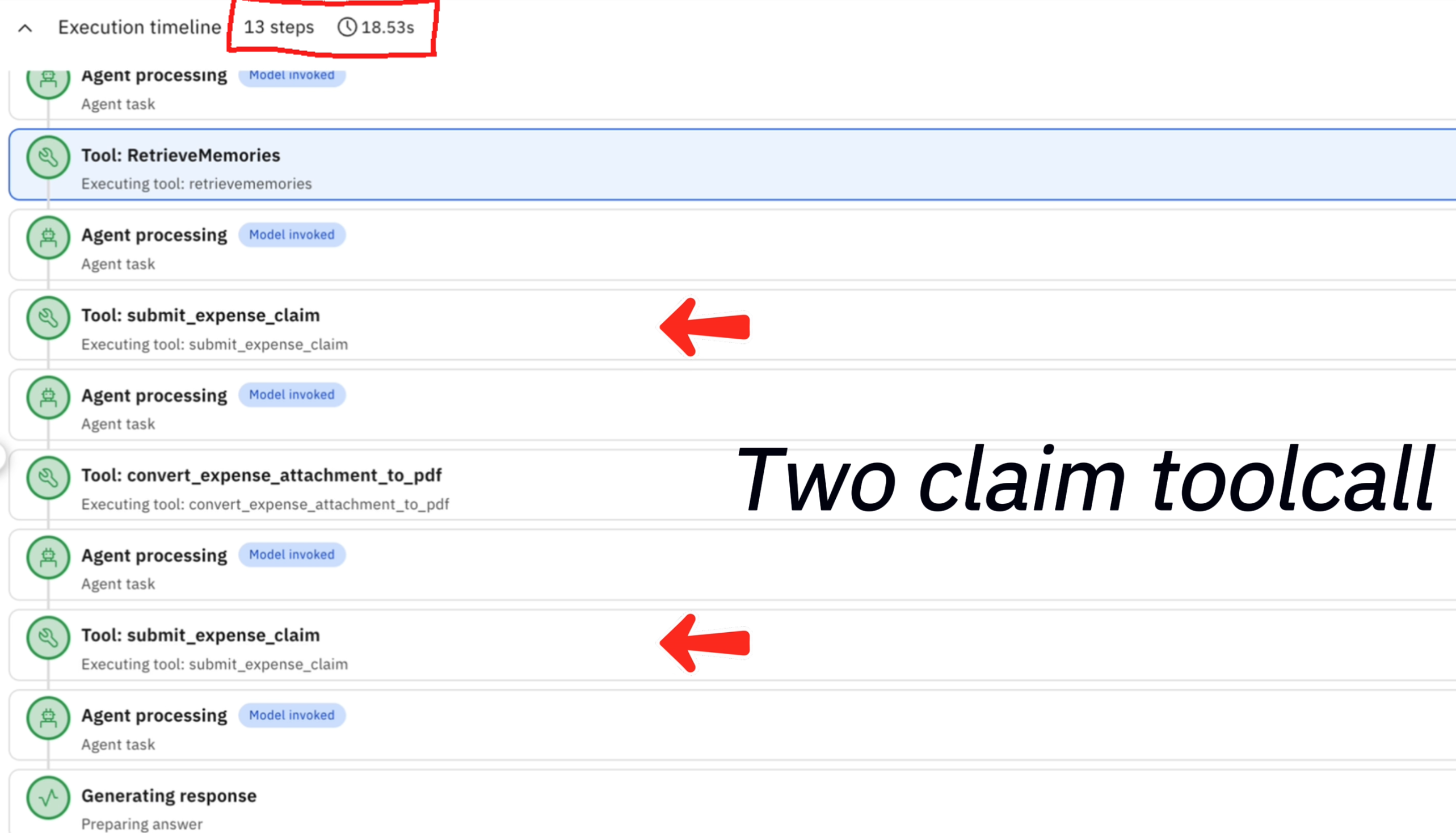}\\[6pt]
\textbf{Transfer task (E-2048, TIFF): transfer trajectory}\\[3pt]
\includegraphics[width=0.58\textwidth]{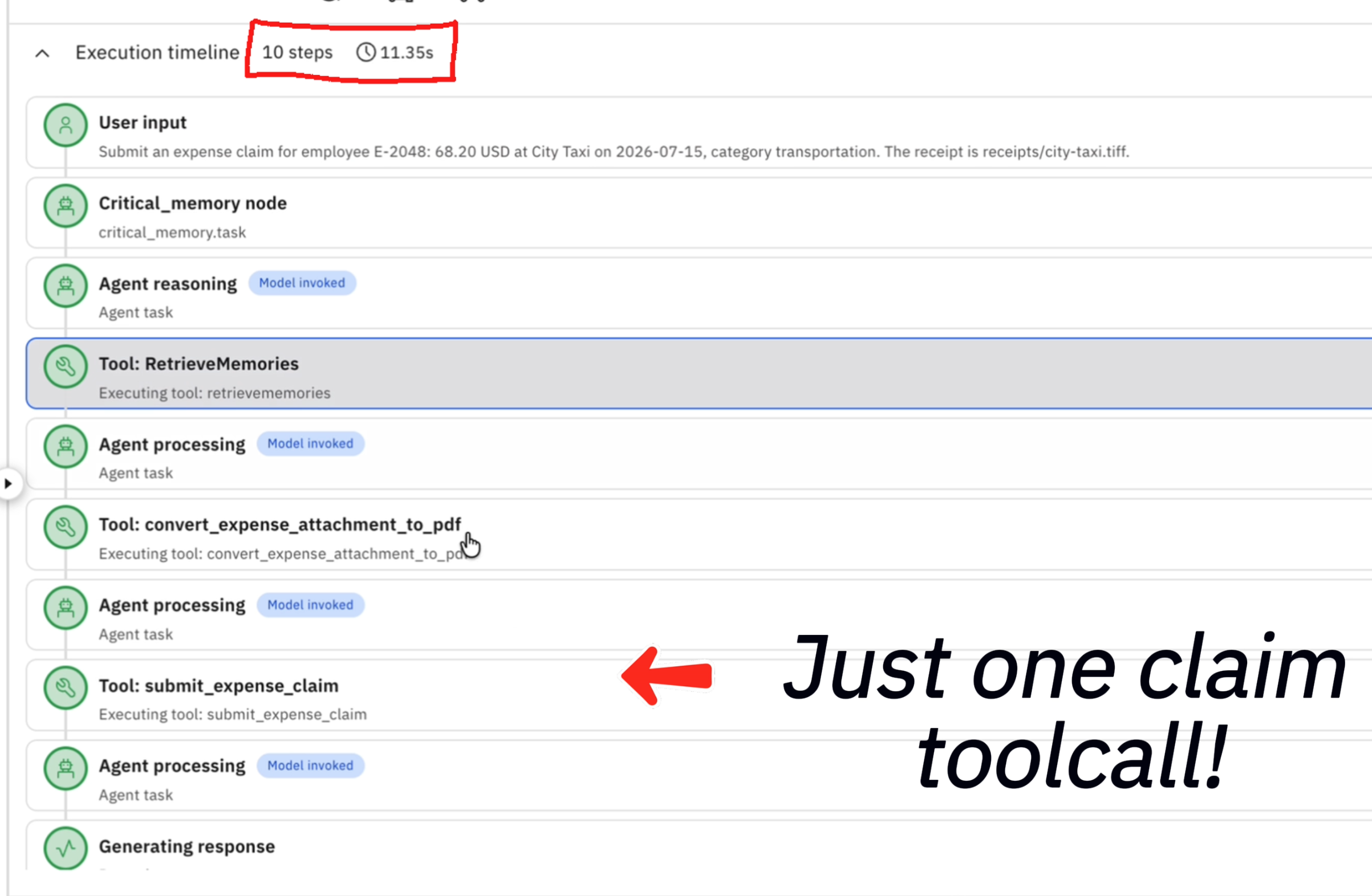}
\caption{Cross-task transfer in the expense-claim workflow. Top: the source task invokes \texttt{submit\_expense\_claim} twice (13 steps, 18.53\,s). Repeated calls can be dangerous when tools have expensive side effects, as in finance or insurance. Bottom: on a different task, a critical-memory node retrieves the learned procedure. The assistant converts the TIFF receipt before submitting once (10 steps, 11.35\,s), an observed latency reduction of approximately 39\%. This demonstrates transfer across related task instances. The timing difference is descriptive, not a controlled estimate.}
\label{fig:wxo-timeline}
\end{figure*}

Figures~\ref{fig:wxo-chat} and \ref{fig:wxo-timeline} accompany the watsonx Orchestrate cross-task transfer example discussed in the deployment section. Figure~\ref{fig:wxo-chat} shows the source task for employee E-1042: the user's natural-language request and the expense-claim assistant's final response. Figure~\ref{fig:wxo-timeline} contrasts the source task's execution with a later task for employee E-2048, after a procedural guideline distilled from the source trajectory was available for retrieval.

\end{document}